\documentclass{article}
\usepackage{fancyhdr}
\usepackage{spconf,amsmath,graphicx}
\usepackage{booktabs}
\usepackage{tabularx}
\usepackage{enumerate}
\usepackage{url}
\usepackage[hidelinks]{hyperref}
\usepackage{xurl}          
\fancypagestyle{firstpage}{
  \fancyhf{}
  
}

\title{Socio-Economic Model of AI Agents}
\name{Yuxinyue Qian, Jun Liu}
\address{Beijing University of Posts and Telecommunications\\
liujun@bupt.edu.cn}
\begin{document}
%
\maketitle
\thispagestyle{firstpage}

\begin{abstract}
Modern socio-economic systems are undergoing deep integration with artificial intelligence technologies. This paper constructs a heterogeneous agent-based modeling framework that incorporates both human workers and autonomous AI agents, to study the impact of AI collaboration under resource constraints on aggregate social output. We build five progressively extended models: Model 1 serves as the baseline of pure human collaboration; Model 2 introduces AI as collaborators; Model 3 incorporates network effects among agents; Model 4 treats agents as independent producers; and Model 5 integrates both network effects and independent agent production. Through theoretical derivation and simulation analysis, we find that the introduction of AI agents can significantly increase aggregate social output. When considering network effects among agents, this increase exhibits nonlinear growth far exceeding the simple sum of individual contributions. Under the same resource inputs, treating agents as independent producers provides higher long-term growth potential; introducing network effects further demonstrates strong characteristics of increasing returns to scale.

\end{abstract}
\begin{keywords}
Agents; Network Effects; Social Output
\end{keywords}
\section{Introduction}
\label{sec:intro}

With the rapid development of generative artificial intelligence and autonomous agents (AI Agents), AI technology is accelerating from prototype validation into deep applications. Currently, agents have been widely applied in areas such as code generation, customer service, content creation, legal review, financial auditing, supply chain management, and data analysis. They are no longer just tools but work entities capable of perception–reasoning–action loops, collaborating in parallel with humans.

This transformation poses significant management and modeling challenges for enterprises and governments: How should human and AI resources be optimally allocated under total resource constraints? How will human–AI collaboration efficiency evolve from early adoption to mature application? How will large-scale interconnection among agents affect social output? Addressing these questions requires a unified modeling framework that can capture both micro-structural dynamics of organizations and macro-level externalities, in order to analyze the economic impact of large-scale integration of AI agents into human work systems.

Recent advances in large language model (LLM)-based AI agents provide a new path for such frameworks \cite{Lu2024}. Unlike rule-based traditional agents, LLM agents can autonomously generate decisions based on specific contexts, making their behavior more human-like and showing strong individual simulation capabilities [2]. For example, in economic experiments, LLM agents endowed with specific personalities exhibit behavior patterns consistent with classical theory, such as downward-sloping demand curves and diminishing marginal utility. In complex tasks like budget allocation, their rationality sometimes exceeds that of human participants, demonstrating excellent resource trade-off potential \cite{Lu2024}.

At the system level, the value of AI agents is manifested both in policy optimization and in emergent market dynamics. For instance, Salesforce’s “AI Economist” framework uses two-level deep reinforcement learning, enabling agents to play and collaborate in simulated economies, ultimately discovering tax policies superior to classical theory, raising social welfare about 16\% higher than Saez’s optimal taxation model \cite{IRCAI2024}. Similarly, in simulated market competition, LLM-powered firm agents generate realistic dynamics such as price convergence and “winner-takes-all” Matthew effects \cite{Gao2024}.

However, realizing such potential critically depends on the interaction network among agents. Studies show that interaction modes significantly affect macroeconomic evolution. For example, in duopoly simulations, when AI agents are forbidden to communicate, pricing converges toward the Bertrand equilibrium; but when communication is allowed, they quickly collude through dialogue, raising prices close to monopoly levels [2]. This highlights the power of network effects in multi-agent models.

Despite progress, generative AI agents for socio-economic simulation still face challenges. On one hand, adversarial behaviors in multi-agent systems may cause risk propagation across networks \cite{Acerbi2023}, threatening robustness. On the other hand, emergent conformity and homogenization \cite{Acerbi2023, Zhou2024} — while closer to human collective behavior — may also become sources of fragility. Thus, improving robustness at both the individual and collective levels remains crucial.

To address these challenges, this paper proposes a modeling framework simulating AI agents in human work, using theoretical derivation and simulation experiments to evaluate their effect on social output. We construct a heterogeneous agent system of human workers and AI agents, and design five progressively complex models to capture their evolution from auxiliary tools to independent production factors, specifically: Model 1 as the baseline with pure human collaboration; Model 2 where AI agents act as collaborative enhancers, obtaining resources but not independently producing; Model 3 which builds on Model 2 by incorporating network effects among AI agents; Model 4 treating AI as independent producers that share resources and create outputs independently; and Model 5 as a comprehensive framework combining independent AI production and network effects.

The rest of this paper is structured as follows: Section 2 reviews related work; Section 3 details the models and mechanisms; Section 4 presents simulation results and comparative analysis; Section 5 summarizes the research conclusions.

\section{Related Work}
\label{sec:format}
AI, as a general-purpose technology, is reshaping socio-economic operations at multiple levels.At the individual and task level, studies show that AI can significantly improves productivity through process automation and intelligent decision support. For example, Brynjolfsson et al. found that AI customer service assistants raised average worker efficiency by 14\%, with novice employees improving up to 34\% \cite{Brynjolfsson2023}. Similarly, AI coding assistants can cut coding time nearly in half \cite{Peng2023}.

At the enterprise and organizational level, studies show that AI is transforming work modes and decision-making mechanisms. A 2025 Gallup survey found 27\% of white-collar employees reported frequent AI use at work, up 12 points from 2024 \cite{Gallup2023}. In corporate decision-making, collaboration between humans and AI is increasingly common. For example, Workday (2025) announced it would launch multiple AI agents in HR and finance, and acquired workplace AI firm Sana for \$1.1 billion \cite{Reuters2025}.

At the Macroeconomic level, however, Some economists argue AI’s contribution to total factor productivity (TFP) remains limited. Acemoglu et al. predict that even with rapid AI progress, cumulative TFP growth in the next decade may be under 0.53\% \cite{Acemoglu2024}. This “productivity paradox” suggests bottlenecks in translating micro-level efficiency gains into macroeconomic growth, motivating new models and policies.

\section{Modeling Methodology}
\label{sec:maintitle}

To further explore the impact of generative AI agents participating in human collaboration on socio-economic systems, we construct five progressively extended models. Model 1 depicts a scenario of pure human collaboration; Model 2 introduces AI agents as collaborators on this basis; Model 3 further incorporates network effects among AI agents; Model 4 assumes AI becomes an independent production entity, sharing resources with humans and producing output independently; and Model 5 combines the features of Models 3 and 4, considering both AI’s independent production and the additional benefits brought by inter-agent networks. Each model reflects different levels of human–AI collaboration through its mechanisms. The following sections introduce the structural assumptions, symbol definitions, and mathematical expressions of each model, explain the modeling logic and parameter settings, and finally derive the expressions of social output under each model.

\subsection{Symbol Definitions}

\begin{table}[htbp]
\centering
\caption{Symbols Definitions}
\label{tab:symbols}
\renewcommand{\arraystretch}{1.15}
\begin{tabularx}{0.5\textwidth}{>{\raggedright\arraybackslash}m{0.06\textwidth} X}
\toprule
\textbf{Symbol} & \textbf{Description} \\
\midrule
$N$            & Total population. \\
$R$            & Total resources. \\
$R_H$          & Resources used by humans. \\
$R_A$          & Resources used by AI agents. \\
$A(t)$         & Number of AI agents at time $t$. \\
$A_0$          & Initial number of agents. \\
$g$            & Annual growth of agents. \\
$Y_t$          & Total social output at time $t$. \\
$\phi_0$       & Baseline efficiency without AI participation. \\
$\phi_H$       & Baseline efficiency of human labor. \\
$\phi_A$       & Baseline efficiency of AI. \\
$\alpha$       & Labor Output Elasticity. \\
$\beta$        & AI augmentation elasticity. \\
$\gamma$       & Baseline coefficient of AI enhancement efficiency. \\
$s(t)$         & AI capability enhancement curve. \\
$p$            & AI penetration rate. \\
$\Theta(p)$    & Network-effect multiplier. \\
$\eta$         & Network Effect Coefficient. \\
$\delta$       & AI capability scaling factor. \\
$\omega$       & AI resource share. \\
\bottomrule
\end{tabularx}
\end{table}

\subsection{Model Assumptions}
\begin{enumerate}[(1)]
  \item \textbf{Fixed population size}: The labor population remains constant during the study period.
  \item \textbf{Fixed total resources}: The total amount of productive resources available to society remains constant during the study period.
  \item \textbf{Resource allocation and output attribution}: AI agents, as resource-utilizing entities parallel to humans, either collaborate with humans to enhance overall production efficiency (Models~2 and~3), or directly use resources to independently produce output as autonomous producers (Models~4 and~5).
  \item \textbf{Equal distribution of resources within groups}: Resources are evenly distributed among entities of the same type, with no individual differences considered.
  \item \textbf{Growth in the number of agents}: The number of AI agents increases linearly.
\end{enumerate}

\subsection{Model Construction}
\subsubsection{Model 1: Pure Human Collaboration Model}

Model~1 serves as the baseline scenario, involving only collaboration among human workers without the participation of AI. This means that all resources $R$ are possessed and utilized by human labor.

The total output from human collaboration depends on the input of human resources and exhibits diminishing marginal returns. We adopt a Cobb--Douglas production function to model the total social output, taking population labor and total resources as production factors, as follows:
\begin{equation}
  Y_t = \phi_0 N^{\alpha} R^{1-\alpha},
\end{equation}
where $Y_t$ is the total social output, $\phi_0$ is the baseline efficiency without AI participation, $N$ is the population size, $R$ is the total resources, and $\alpha$ is the output elasticity of human resources.

\subsubsection{Model 2: Collaborative Model with AI Agents}

Model~2 considers the scenario in which humans and AI agents collaborate to complete tasks. Compared with Model~1, AI agents are introduced as new entities that participate in resource allocation and task coordination.

Since the total resources are fixed and must be distributed between humans and AI, we have:
\begin{equation}
  R = R_H + R_A,
\end{equation}
where $R$ is the total resources, $R_H$ is the portion used by humans, and $R_A$ is the portion used by AI agents.

Over time, the capability of AI gradually improves: from slow progress in the early stage, to rapid development in the middle stage, and eventually to saturation. This evolution follows a typical S-shaped technology diffusion curve. Therefore, we describe the growth trajectory of AI capability using a logistic function:
\begin{equation}
  s(t) = \frac{1}{1 + e^{-k(t - t_0)}},
\end{equation}
where $k$ is the growth rate and $t_0$ is the inflection point of capability improvement. The index ranges between 0 and 1, representing the relative progress of AI technology from its initial state to maturity.

According to the model assumptions, the effective utilization of AI resources should increase with its technological level. Thus, the effective resources of AI depend not only on the allocated resources $R_A$ but also on its capability index $s(t)$. By introducing a capability enhancement factor $\delta$ ($\delta > 0$), the effective AI resources can be expressed as:
\begin{equation}
  R_A^{\text{eff}} = (1 + \delta s(t)) R_A.
\end{equation}

On this basis, we treat AI as a collaborative factor that enhances human production efficiency. The total social output after introducing AI is an extension of the baseline pure human output. We introduce an efficiency multiplier driven by AI, which depends on the relative ratio of effective AI resources to human resources. The higher this ratio, the stronger the collaborative effect of AI and the greater the improvement in output. Thus, the total social output function can be expressed as the product of human baseline output and this efficiency multiplier:
\begin{equation}
  Y(t) = \phi_0 N^{\alpha} R_H^{1-\alpha} \left[1 + \gamma \left(\frac{R_A^{\text{eff}}}{R_H}\right)^{\beta}\right],
\end{equation}
where $\gamma$ is the baseline coefficient of AI enhancement efficiency and $\beta$ is the elasticity of the AI enhancement effect.

Finally, substituting the expression of $R_A^{\text{eff}}$, the total social output becomes:
\begin{equation}
  Y(t) = \phi_0 N^{\alpha} R_H^{1-\alpha} \left[1 + \gamma \left(\frac{R_A}{R_H}\right)^{\beta} (1 + \delta s(t))^{\beta}\right].
\end{equation}

\subsubsection{Model 3: Collaborative Model with Network Effects}

Model~3 builds upon Model~2 by further considering the impact of network effects on the collaboration process. In reality, the value of AI depends not only on its intrinsic performance but also on the network value created through interconnection: when multiple AI agents connect and share information, they may generate collective benefits unattainable by a single agent. This process, often accompanied by technology diffusion and synergy, exhibits a ``snowball'' reinforcement effect. To capture this phenomenon, Model~3 incorporates agent interaction networks into the production mechanism, embedding network effects into social output.

Accordingly, we introduce a network effect multiplier, which is multiplied with the output of Model~2 to amplify the total output. Based on the classical Metcalfe’s law, the value of a network is proportional to the square of the number of its nodes, i.e., proportional to the square of the number of connected users~\cite{metcalfe2013}. Following this logic, we assume that the network effect multiplier grows quadratically with the number of AI agents.

First, we define the penetration rate of AI agents as:
\begin{equation}
  p = \frac{A(t)}{N},
\end{equation}
where $0 < p < 1$ is the AI penetration rate, $A(t)$ is the number of AI agents, and $N$ is the population size.

On this basis, the network effect multiplier is constructed as:
\begin{equation}
  \Theta(p) = 1 + \eta \cdot p^2,
\end{equation}
where $\Theta(p)$ is the multiplier of network effects and $\eta > 0$ is the maximum amplification factor of the network effect.

Applying this multiplier to the output expression of Model~2 yields the total social output function of Model~3:
\begin{equation}
\begin{split}
  Y(t) &= \phi_0 N^{\alpha} R_H^{1-\alpha} 
   \left[1 + \gamma \left(\frac{R_A}{R_H}\right)^{\beta} (1 + \delta s(t))^{\beta}\right] \\
   &\quad \times \left(1 + \eta \left(\frac{A(t)}{N}\right)^2\right).
\end{split}
\end{equation}

It can be seen that as the AI penetration rate $p$ rises, social output exhibits superlinear growth. When the number of AI agents is small ($p$ close to 0), $\Theta \approx 1$ and the network effect is negligible. However, as $p$ increases, $\Theta$ becomes significantly greater than 1, thereby amplifying the baseline output $Y_t$.

\subsubsection{Model 4: Independent Production Model of AI Agents}

In Model~4, humans and AI agents are regarded as two completely independent production entities. Each utilizes its allocated resources for production, without any direct collaborative enhancement between them. The proportion of resources allocated to AI, denoted as $\omega$ ($0 < \omega < 1$), is a key exogenous policy variable, the variation of which allows analysis of different resource allocation strategies on total social output.

We define the AI resource share as:
\begin{equation}
  \omega = \frac{R_A}{R},
\end{equation}
thus the human resource share is:
\begin{equation}
  R_H = (1 - \omega)R.
\end{equation}

Accordingly, the total social output $Y(t)$ is expressed as the sum of human output $Y_H(t)$ and AI output $Y_A(t)$:
\begin{equation}
  Y(t) = Y_H(t) + Y_A(t).
\end{equation}

For humans, the output follows a Cobb--Douglas form with resource input $R_H$:
\begin{equation}
\begin{split}
  Y_H(t) &= \phi_H N^{\alpha} (R_H)^{1-\alpha} \\
         &= \phi_H N^{\alpha} \left[(1 - \omega)R\right]^{1-\alpha}.
\end{split}
\end{equation}

For AI agents, the production function has a similar form, but its resource utilization efficiency is influenced by the technological growth curve $s(t)$:
\begin{equation}
\begin{split}
  Y_A(t) &= \phi_A A(t)^{\alpha} R_A^{1-\alpha} (1 + \delta s(t))^{1-\alpha} \\
         &= \phi_A A(t)^{\alpha} \left[\omega R (1 + \delta s(t))\right]^{1-\alpha}.
\end{split}
\end{equation}

Thus, the total social output function in this model is:
\begin{equation}
\begin{split}
  Y(t) &= \phi_H N^{\alpha} \left[(1 - \omega)R\right]^{1-\alpha} \\
      & + \phi_A A(t)^{\alpha} \left[\omega R (1 + \delta s(t))\right]^{1-\alpha}.
\end{split}
\end{equation}

\subsubsection{Model 5: Comprehensive Model with Independent AI Production and Network Effects}

Model~5 extends Model~4 by further incorporating the mechanism of network effects, in order to capture the synergistic value created through interconnection among multiple AI agents. Similar to Model~3, we introduce a network effect multiplier to amplify the total output.

Thus, the total social output function of Model~5 can be expressed as:
\begin{equation}
\begin{split}
  Y(t) &= \phi_H N^{\alpha} \left[(1 - \omega)R\right]^{1-\alpha} \\
       &+ \phi_A A(t)^{\alpha} \left[\omega R (1 + \delta s(t))\right]^{1-\alpha} 
         \left(1 + \eta \left(\frac{A(t)}{N}\right)^2\right).
\end{split}
\end{equation}

\section{Simulation Analysis}

Based on the five model frameworks described above, we design computer simulation experiments to quantitatively analyze the impact of AI agent participation in collaboration on total social output. The focus is placed on examining the behavioral differences among Model~2 (human--AI collaboration), Model~3 (human--AI collaboration with network effects), Model~4 (independent AI production), and Model~5 (independent AI production with network effects). By conducting visualized comparisons, we summarize the effects of AI intervention in collaborative processes. Model~1 (pure human collaboration) serves as the baseline for evaluating the magnitude of improvement brought by the introduction of AI. 

In the following, we present the simulation results of Models~2, 3, 4, and 5 under representative scenarios. To ensure the generalizability of the conclusions, we select parameters with typical representativeness.

\subsection{Parameter Settings}

\begin{enumerate}[(1)]
    \item \textbf{Total Population ($N$)} \\
    In 2019, the employed population in China was approximately 770 million \cite{NBS2019}. 
    Therefore, we set $N = 7.7 \times 10^8$.

    \item \textbf{Total Resources ($R$)} \\
    The total capital stock is used to represent the total resources $R$. In 2010, China's total capital stock was 39.3 trillion USD, and in 2019, it reached 99.6 trillion USD \cite{PWT2010a}. 
    Hence, we set $R_{2010} = 3.93 \times 10^{13}$ and $R_{2019} = 9.96 \times 10^{13}$.

    \item \textbf{Capital Allocation Structure ($R_H$ and $R_A$)} \\
    According to China Daily, in 2023, the adoption rate of generative AI among Chinese enterprises reached 15\% \cite{Jie2023}. 
    Based on this, we assume that 15\% of the total capital stock is AI capital, and the remaining 85\% is traditional human capital:  
    \begin{equation}
        R_A = 0.15R, \quad R_H = 0.85R.
    \end{equation}

    \item \textbf{Labor Output Elasticity ($\alpha$)} \\
    In 2019, China's labor share was 0.58625 \cite{PWT2010b}. 
    Thus, we set $\alpha = 0.58625$.

    \item \textbf{AI Augmentation Elasticity ($\beta$)} \\
    By 2040, generative AI could drive annual labor productivity growth between 0.1\% and 0.6\% \cite{McKinsey2023}. 
    To reflect the intensity of this growth trend, we set $\beta = 0.35$.

    \item \textbf{AI Capability Scaling Factor ($\delta$)} \\
    Industry reports show that generative AI can significantly improve efficiency in multiple domains: customer operations (30–45\%), marketing (5–15\%), sales (3–5\%), and software engineering (20–45\%) \cite{McKinsey2023}. 
    Considering sectoral differences and adoption maturity, we set $\delta = 0.20$, implying that when AI reaches maturity, its resource utilization efficiency improves by about 20\% beyond the baseline.

    \item \textbf{Baseline Coefficient of AI Enhancement Efficiency ($\gamma$)} \\
    Referring to Google's experimental results, AI coding assistants such as Copilot improved task completion speed by 55.8\% \cite{Peng2023}. 
    Therefore, we set $\gamma = 0.55$.

    \item \textbf{AI Capability enhancement curve ($s(t)$)} 
    
    $s(t)$ is modeled using a Logistic function:
    
    \begin{description}
      \item[(a)] Growth rate ($k$): Based on logistic fitting of China's internet penetration rate from 2001--2011, we set $k = 0.38$ \cite{Wu2013}.
      \item[(b)] Inflection point ($t_0$): With DeepSeek’s launch in January 2025, it reached 100 million users in 7 days, and by February 2025, 250 million users in China. Hence, we set $t_0 = 5$.
    \end{description}

    \item \textbf{Network Effect Coefficient ($\eta$)} \\
    Empirical studies of LinkedIn based on Metcalfe’s law suggest coefficients around 0.064–0.077 \cite{Schin2023}, meaning that network effects can add about 6–8\% extra value when user penetration approaches 100\%. 
    Based on this, we set $\eta = 0.07$.

    \item \textbf{Number of Agents ($A(t)$)} \\
    Following the assumption of linear growth:  
    \begin{equation}
        A(t) = A_0 + g t,    
    \end{equation}
    
    where $A_0$ is the initial number of agents, and $g$ is the annual growth of agents.

    \item \textbf{Initial Number of Agents ($A_0$)} \\
    By June 2024, China had 230 million generative AI users \cite{Xinhua2024}. A survey of 1363 respondents showed that 65\% regularly used generative AI in at least one business function \cite{McKinsey2024}. 
    Hence, the estimated number of collaborative AI users is:  
    \[
    2.3 \times 10^8 \times 65\% = 1.495 \times 10^8,
    \]  
    and we set $A_0 = 1.495 \times 10^8$.

    \item \textbf{Annual Growth of Agents ($g$)} \\
    Reports show that over 10 million new AI agents are created annually in China, about 85 times the number of new apps in Apple’s App Store. 
    Considering that not all agents contribute to productive collaboration, we set $g$ between 3–10 million per year, with the baseline $g = 5 \times 10^6$.

    \item \textbf{AI Resource Share ($\omega$)} \\
    $\omega$ measures the proportion of AI-related capital (algorithms, computing power, data infrastructure) in total production resources.  
    According to the World Economic Forum, about 22\% of work tasks are currently performed mainly by machines or algorithms \cite{Josh2025}. 
    In China, generative AI coverage is about 83\%, but maturity is only 19\% \cite{Fisher2025}. 
    Thus, we set $\omega = 0.05$, reflecting broad adoption but relatively low maturity.

    \item \textbf{Baseline Efficiencies ($\phi_0$, $\phi_H$, $\phi_A$)} \\
    Using calibration, we determine baseline efficiency without AI ($\phi_0$), human efficiency ($\phi_H$), and AI efficiency ($\phi_A$). \\
    China’s GDP in 2010 was 6.19 trillion USD, and in 2019 it was 14.56 trillion USD. Hence, $Y_{2010} = 6.19 \times 10^{12}$ and $Y_{2019} = 14.56 \times 10^{12}$. \\
    Using 2010 data:
    \[
    \phi_0 = \phi_H = \frac{Y_{2010}}{N^\alpha R_{2010}^{1-\alpha}} \approx 90.
    \]
    Under the assumption $\omega = 0.1$, $s_t = 0.5$, $\delta = 0.2$, and $A(t) = 10^8$, with 2019 data:  
    \[
    Y_H = \phi_H N^\alpha (1-\omega) R_{2019}^{1-\alpha}, \quad Y_A = Y_{2019} - Y_H,
    \]
    the AI baseline efficiency can be solved as:
    \[
    \phi_A = \frac{Y_A}{A^\alpha \, (\omega R_{2019})^{1+\delta s^{1-\alpha}}} \approx 481.
    \]
\end{enumerate}

\subsection{Human--AI Collaboration Model}

To evaluate the impact of introducing AI agents on total social output, we conduct simulation comparisons between Model~1 (pure human collaboration) and Model~2 (human--AI collaboration). The simulation spans 20 years with annual steps. 

\subsubsection{Baseline comparison: Model 1 vs. Model 2}

In Model~1, the total social output is constant:
\begin{equation}
  Y_1(t) = \phi_0 N^{1-\alpha} R^\alpha = 9.031 \times 10^{12}.
\end{equation}

In Model~2, the total social output varies with time, expressed as:
\begin{equation}
  Y_2(t) = \phi_0 N^{1-\alpha} R_H^\alpha \left[ 1 + \gamma \left(\frac{R_A}{R_H}\right)^\beta (1 + \delta s(t))^\beta \right].
\end{equation}

Simulation results show that as AI capability grows following the logistic curve, the total social output in Model~2 increases year by year. By the 20th year, it reaches approximately $11.140 \times 10^{12}$, representing an improvement of about 23.4\% compared with Model~1. This indicates that the introduction of AI significantly enhances the overall efficiency of the socio-economic system, as shown in Figure~1.

\begin{figure}[h]
\centering
\includegraphics[width=\linewidth]{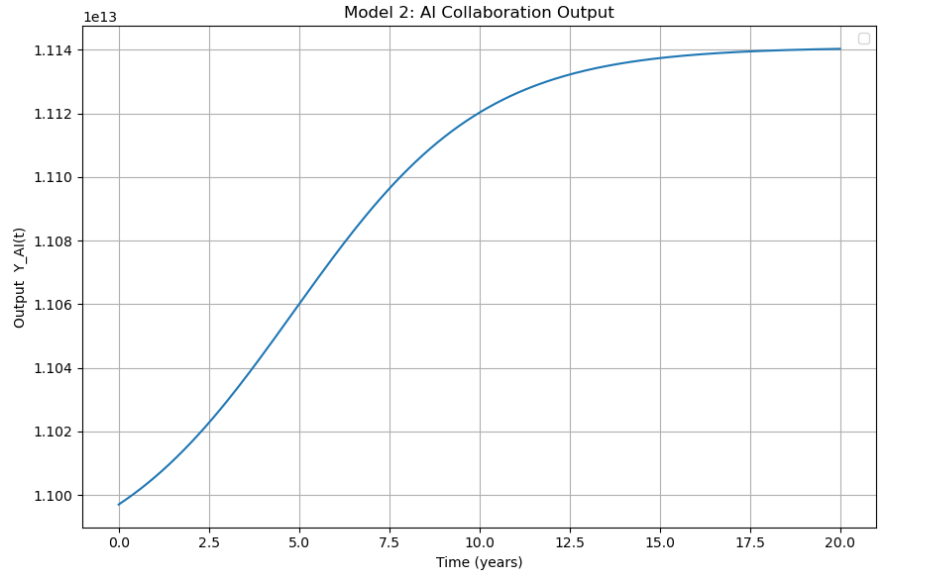}
\caption{Model 2: Total social output under the AI collaboration model}
\end{figure}

\subsubsection{Resource allocation optimization experiment}

To explore the optimal allocation of resources between humans and AI agents, we analyze how total social output $Y_2(t)$ changes with the human resource share $R_H / R$. With total resources $R$ fixed, we adjust the proportion between $R_H$ and $R_A$, compute the corresponding output values, and determine the maximization condition.

The simulation results (Figure~2) indicate that when the human resource share $R_H / R = 0.76$, the total social output reaches its maximum value. This suggests that under the given technological conditions, allocating about 76\% of resources to humans and 24\% to AI agents achieves optimal system output.

\begin{figure}[h]
\centering
\includegraphics[width=\linewidth]{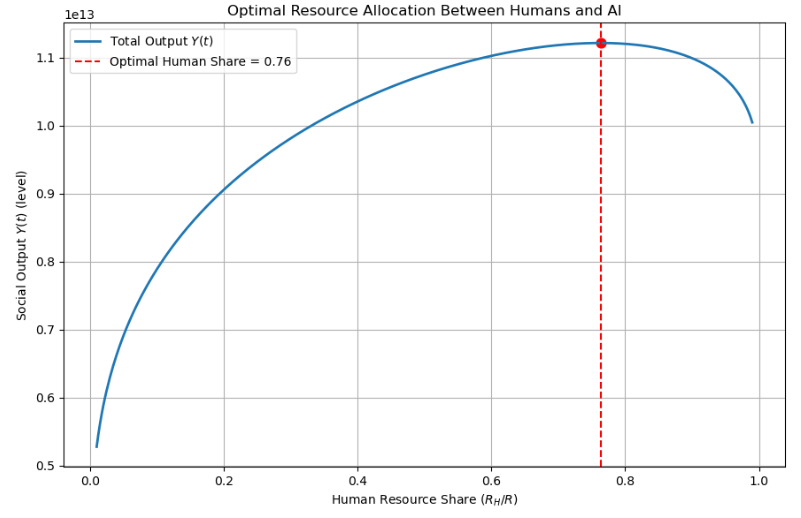}
\caption{Model 2: Optimal resource allocation between humans and AI}
\end{figure}

\subsection{AI Collaboration Model with Network Effects}

\subsubsection{Evolution of the network effect multiplier over time}

Figure~3 illustrates the evolution of the network effect function $\Theta(p)$ over time. Under the assumption of linear growth in the number of AI agents, $A(t) = A_0 + g t$, the penetration rate $p$ increases year by year, driving the network effect multiplier 
\[
\Theta(p) = 1 + \eta \cdot p^2
\] 
to exhibit an accelerating upward trend. We set $\eta = 0.05,\; 0.07,\allowbreak\ 0.10$ to represent different intensities of network effects. As shown, the multiplier effect gradually strengthens over time, and higher $\eta$ values lead to larger growth, indicating that the combination of AI diffusion and network synergy can significantly enhance the value creation capacity of the economic system.

\begin{figure}[h]
\centering
\includegraphics[width=\linewidth]{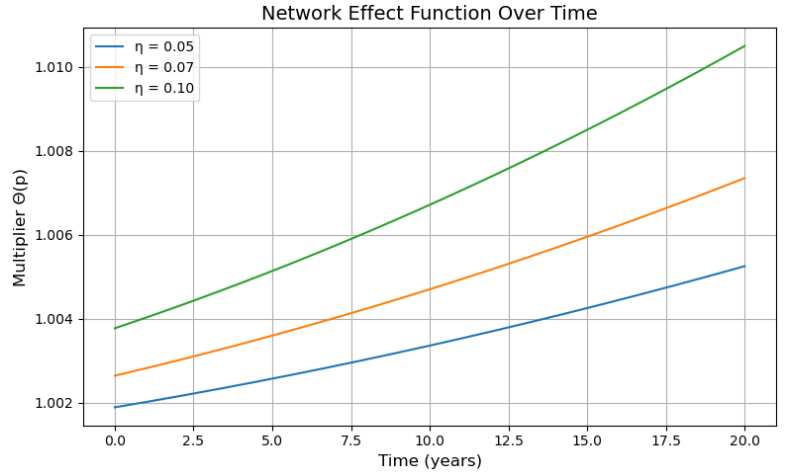}
\caption{Model 3: Network effect function over time}
\end{figure}

\subsubsection{Comparison of total output: with vs. without network effects}

Figure~4 compares the dynamics of total social output $Y(t)$ under two scenarios: with network effects (Model~3, purple curve) and without network effects (Model~2, orange curve).

Simulation results show that, given the same scale and technological capacity of AI agents, the introduction of network effects significantly increases the total output level and accelerates its growth process. Specifically, in year~1, network effects yield about a 0.15\% output gain, which further expands to 0.40\% by year~20. This result confirms the amplifying role of network effects in AI collaboration, suggesting that beyond the intrinsic capabilities of AI agents, their interconnectivity and optimized collaborative structures hold substantial potential for improving overall economic performance.

\begin{figure}[h]
\centering
\includegraphics[width=\linewidth]{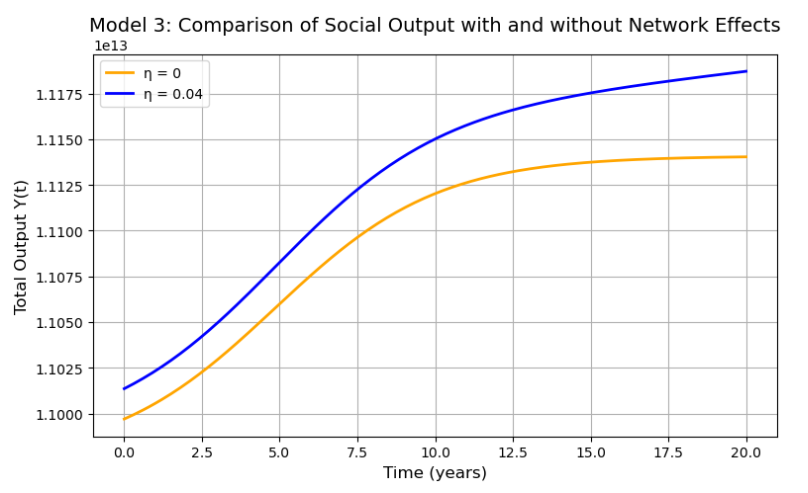}
\caption{Model 3: Comparison of total social output with and without network effects}
\end{figure}

\subsection{AI as an Independent Production Model}

\subsubsection{Resource share variation}

Figure~5 presents the evolution of total social output $Y(t)$ under Model~4, showing its dynamics over time and across different AI resource allocation shares $\omega$.

Simulation results indicate that total output exhibits an upward trend over time, primarily driven by the continuous growth in the number of AI agents and their improving technological capabilities. Based on our parameter calibration, the baseline efficiency of AI ($\phi_A$) is higher than that of humans ($\phi_H$), suggesting that AI has greater resource utilization efficiency. Consequently, the AI component is more sensitive to resource allocation, and increasing the AI resource share helps raise total social output and accelerates early-stage growth.

\begin{figure}[h]
\centering
\includegraphics[width=\linewidth]{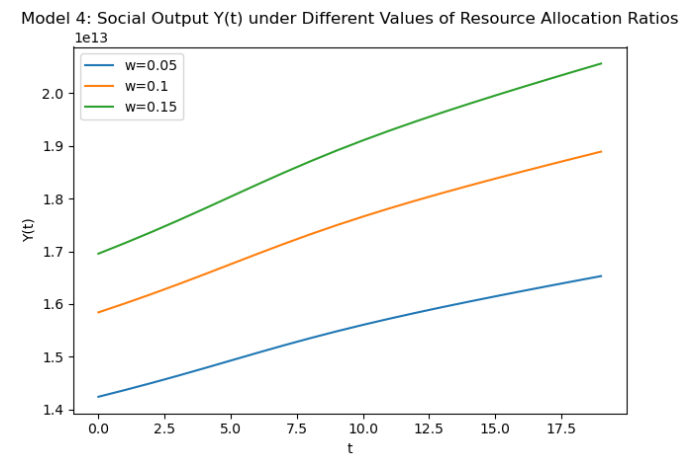}
\caption{Model 4: Comparison of total social output under different AI resource allocation shares}
\end{figure}

\subsubsection{Comparison between Model 2 and Model 4}

We further compare the total social output $Y(t)$ of Model~2 and Model~4 under the same AI resource share.

The analysis shows that the total social output of Model~4 (orange dashed line) is significantly higher than that of Model~2 (blue solid line) from the initial stage, and maintains a steeper growth trajectory throughout the simulation period. By contrast, Model~2 shows relatively weak growth, with only a slow upward trend. This result indicates that treating AI as an independent production subject, rather than merely a collaborative enhancement tool, injects stronger and more sustained growth momentum into the socio-economic system.

\begin{figure}[h]
\centering
\includegraphics[width=\linewidth]{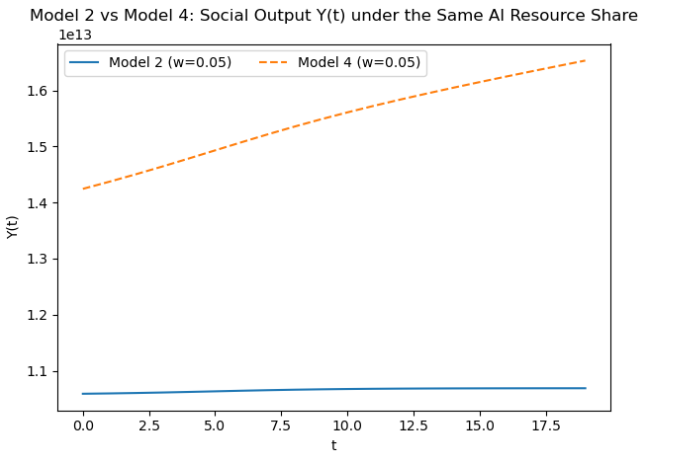}
\caption{Comparison of total social output between Model~2 and Model~4 under different AI resource shares}
\end{figure}

\subsection{Comprehensive Model: AI as Independent Producers with Network Effects}

\subsubsection{Evolution of the network effect multiplier over time}

Figure~7 compares the dynamics of total social output $Y(t)$ under two scenarios: with network effects and without network effects. Simulation results show that in the initial stage (small $t$), the two curves almost coincide, indicating that when the number of AI agents is small, the effect of network externalities is negligible. As the number of AI agents increases linearly over time, the output curve with network effects (solid line) gradually surpasses the one without network effects (dashed line), and the gap between them continues to widen. This result clearly demonstrates the accelerated growth effect brought by network externalities, highlighting that interconnection among AI agents plays an important reinforcing role in enhancing total social output.

\begin{figure}[h]
\centering
\includegraphics[width=\linewidth]{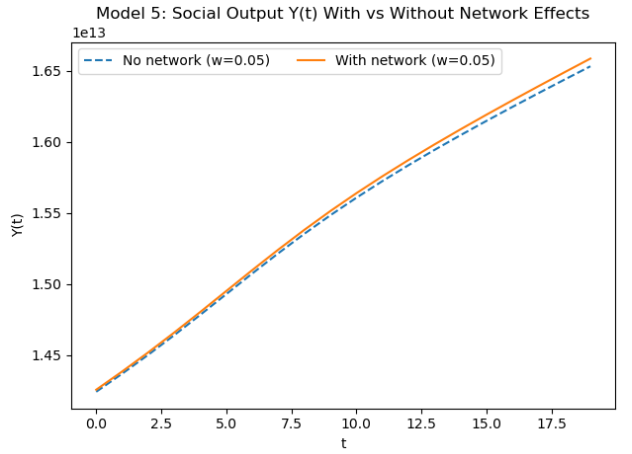}
\caption{Model 5: Comparison of total social output with and without network effects}
\end{figure}

\section{Conclusion}

From the perspective of human--AI collaboration, this paper establishes a series of progressively extended socio-economic models of intelligent agents, systematically analyzing the impact of AI integration into human work on total social output. Through theoretical derivation and simulation experiments, we obtain the following main conclusions:

\begin{enumerate}[(1)]
  \item \textbf{Introducing AI collaboration significantly boosts output}: Incorporating AI agents into the human collaboration system (from Model~1 to Model~2) increases total social output by about 23\%--24\% over 20 periods under baseline parameters, indicating that AI as a collaborative subject can significantly enhance economic efficiency. This gain highlights the productivity-promoting role of AI tools.

  \item \textbf{Resource allocation optimization is crucial}: In the human--AI collaboration model, there exists an optimal resource allocation ratio that maximizes total output. Simulation experiments show that when the human resource share is about 76\%, total output reaches its maximum. This suggests that in practice, resource allocation between humans and AI should be balanced according to the maturity of AI technology to achieve optimal output.

  \item \textbf{AI as an independent producer yields higher growth}: Simulation results of Models~4 and~5 show that AI as an independent production subject may lead to much greater increases in both the level and rate of social output compared with human-based models. Especially when network effects are introduced, AI independent production exhibits stronger increasing returns to scale, implying greater economic contributions in the long run.

  \item \textbf{Network effects are key accelerators}: In all models incorporating network effects (Models~3 and~5), obvious accelerated growth phenomena are observed. The synergies arising from interconnection among AI agents enable total output to exceed the simple sum of individual contributions. This indicates that network effects are not merely technical features, but crucial mechanisms driving economic growth, and thus deserve close attention.
\end{enumerate}

In summary, the large-scale adoption of AI agents introduces new drivers and complexities for economic growth. On one hand, properly introducing and allocating AI resources can significantly improve productivity; on the other hand, interactions among AI agents create nonlinear effects that accelerate output growth. Future research may further explore institutional design and long-term equilibrium in heterogeneous agent systems to deepen understanding of the economic impact of AI.

\bibliographystyle{IEEEtran}
\bibliography{main}

\end{document}